\documentclass{article}

\usepackage{arxiv}

\usepackage[utf8]{inputenc} % allow utf-8 input
\usepackage[T1]{fontenc}    % use 8-bit T1 fonts
\usepackage{hyperref}       % hyperlinks
\usepackage{url}            % simple URL typesetting
\usepackage{booktabs}       % professional-quality tables
\usepackage{amsfonts}       % blackboard math symbols
\usepackage{nicefrac}       % compact symbols for 1/2, etc.
\usepackage{microtype}      % microtypography
\usepackage{lipsum}		% Can be removed after putting your text content
\usepackage{graphicx}
\usepackage[square,numbers]{natbib}
\usepackage{doi}
\usepackage{amsthm}
\usepackage{tabularx}
\usepackage{makecell}
\usepackage{pifont}
\usepackage{amsmath,amssymb,amsfonts}
\usepackage{algorithm}
\usepackage{algorithmic}
\usepackage{graphicx}
\usepackage{textcomp}
\usepackage{xcolor}
\DeclareMathOperator{\class}{class}
\DeclareMathOperator{\sign}{sign}
\DeclareMathOperator{\LDA}{LDA}
\DeclareMathOperator{\softmax}{softmax}
\DeclareMathOperator{\prediction}{prediction}
\DeclareMathOperator{\argmax}{argmax}

\newcommand{\cmark}{\ding{51}}%
\newcommand{\xmark}{\ding{55}}%

\title{Deep Residual Compensation Convolutional Network without Backpropagation}

\author{ \href{https://orcid.org/0000-0002-3936-240X}{\includegraphics[scale=0.06]{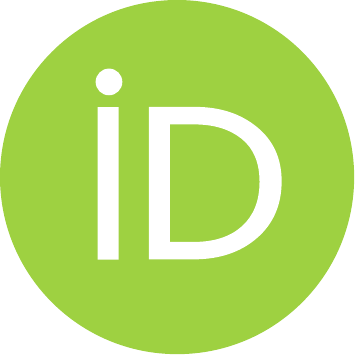}\hspace{1mm}Mubarakah M.~Alotaibi} \\
	Department of Computer Science\\
	University of York, Taif University\\
	York, UK, Taif, Saudi Arabia \\
	\texttt{mmma512@york.ac.uk} \\
	\And
	\href{https://orcid.org/0000-0001-7265-3033}{\includegraphics[scale=0.06]{orcid.pdf}\hspace{1mm}Richard C.~Wilson} \\
	Department of Computer Science\\
	University of York\\
	York, UK\\
	\texttt{richard.wilson@york.ac.uk} \\
}

\renewcommand{\shorttitle}{\textit{arXiv} Template}

\hypersetup{
pdftitle={Deep Residual Compensation Convolutional Network without Backpropagation},
pdfsubject={Deep ResCNet},
pdfauthor={Mubarakah Alotaibi, Richard Wilson},
pdfkeywords={PCANet, DCTNet, DCCNet, CCANet, OSNet, LDANet, Multi-layer PCANet, Image Classification},
}

\begin{document}
\maketitle

\begin{abstract}
PCANet and its variants provided good accuracy results for classification tasks. However, despite the importance of network depth in achieving good classification accuracy, these networks were trained with a maximum of nine layers. In this paper, we introduce a residual compensation convolutional network, which is the first PCANet-like network trained with hundreds of layers while improving classification accuracy. The design of the proposed network consists of several convolutional layers, each followed by post-processing steps and a classifier. To correct the classification errors and significantly increase the network's depth, we train each layer with new labels derived from the residual information of all its preceding layers. This learning mechanism is accomplished by traversing the network's layers in a single forward pass without backpropagation or gradient computations. Our experiments on four distinct classification benchmarks (MNIST, CIFAR-10, CIFAR-100, and TinyImageNet) show that our deep network outperforms all existing PCANet-like networks and is competitive with several traditional gradient-based models. 
\end{abstract}

% keywords can be removed

\section{INTRODUCTION}
\subsection{Background and Related Works}
Deep learning is a non-task-specific learning technique that uses hierarchical structures to automatically learn representations from raw data \citep{ref37}. Since Alexnet \citep{ref2} won the 2012 ImageNet challenge, convolutional neural networks (CNNs) have been widely used for image classification with great success. VGG \citep{ref3}, ResNet \citep{ref4} and Network in network \citep{lin2013network} are a few examples of standard CNNs. Despite their success, CNNs are usually trained with a large number of parameters, requiring intensive parameter updates and a lot of data for training, which might increase the computational cost even with GPU-equipped computing resources \citep{gatto2020semi}. 

\citet{chan2015pcanet} presented PCANet as an alternative to deep CNNs for classification tasks on small datasets. The network structure comprises two cascaded principal component analysis (PCA) layers followed by binary hashing, block-wise histogram and a classifier. Unlike CNNs, PCANet learns its filter bank in a non-iterative layer-by-layer fashion through PCA applied to image-based patches. This training mechanism provides a faster training time advantage to PCANet over conventional CNNs. With its simple architecture, PCANet has achieved state-of-the-art performance across several datasets, including MNIST, extended Yale B, AR and FERET.

The success of PCANet has inspired a family of related strategies. Generally speaking, PCANet-related research has fallen into one of two categories: those focusing on improving the features representation process or those attempting to increase network depth. The work in this paper fits within the second category. 

In an effort to enhance PCANet's feature representation, several articles studied networks that maintained PCANet's fundamental structure while producing different features using different convolutional filters. DCTNet \citep{ng2015dctnet}, CCANet \citep{yang2017canonical} and ICANet \citep{zhang2018icanet} are examples of PCANet-like networks that create their filter banks using unsupervised approaches. The LDANet \citep{chan2015pcanet}, DCCNet \citep{gatto2017discriminative} and OSNet \citep{gatto2017subspace} are a few examples of those that use supervised approaches to produce their filter banks. DFSNet \citep{gatto2020semi} is a good example of a network that uses semi-supervised filters. 

In order to increase the network depth, several studies have attempted to address PCANet's primary issue, the features explosion problem, which limits its depth to only two layers. The block-wise histogram is one factor that contributes to this problem, as the number of bins required to calculate the histogram features grows exponentially with the number of filters in the second stage. \citet{ref40} significantly reduced PCANet's features by replacing the histogram pooling with second-order pooling. The per-channel convolution mechanism is another factor that contributes to the increasing dimensionality problem in PCANet. PCANet+ \citep{ref39} provided an alternative solution to this issue by proposing a filter ensemble mechanism that aggregates the feature maps over all channels, similar to CNNs. By adopting CNN-like filters, as in \citep{ref39}, replacing PCANet's binarisation step with the z-score method and using second-order pooling and late fusion, \citet{alotaibi2021multi} were able to expand PCANet depth to nine layers and outperform the original design. In contrast to prior studies that aimed to increase network depth by tackling the dimensionality issue, our objective is to explore the design of a hundreds-layer PCANet-like network by optimising the classification process at each layer.
\subsection{Proposed Network}
In this article, we present a residual compensation convolutional learning framework to achieve accuracy from a considerably increased depth while simultaneously correcting network errors as we traverse it. The proposed network inherits the simplicity of PCANet-like networks and is trained in a single forward-pass without gradient computations or backpropagation. A comprehensive description of the network architecture, along with its training procedures, is described in Section \ref{architecture}. The experimental section (\ref{exp}) consists of three subsections. In Section \ref{s1}, we evaluate the performance of the proposed network on MNIST, CIFAR-10, CIFAR-100 and TinyImageNet against gradient-based and non-gradient architectures without data augmentation. Section  \ref{ss2} examines the influence of several network parameters, such as the number of filters (\ref{experiment1}) and the learning rate (\ref{expeiment2}), on the accuracy of the proposed model. The final subsection (Section \ref{s2}) addresses the use of data augmentation to improve the model's accuracy. Finally, the conclusions and future works are discussed in Section \ref{s3}.
\section{NETWORK ARCHITECTURE}
\label{architecture}
\subsection{Problem Formulation}
Consider a classification problem with $N$ training images $X^{(1)} \in \mathbb{R}^{m \times n \times d \times N}$, where $m$ and $n$ are the images' spatial dimensions, and $d \in \{1,3\}$ is the number of channels for greyscale or colour images. Let $T$ represent the $C$ classes to which the images originally belonged. The objective of our deep residual compensation convolution network (ResCNet) is to construct a PCANet-like network that achieves high accuracy from a considerable network depth and is trained without using gradient descent or backpropagation. Precisely, the network structure should be structured hierarchically such that, as network depth increases, succeeding layers compensate for the classification errors of previous layers. By combining
the predicted probabilities of the network layers, the model
should obtain a high accuracy.
\subsection{Design Overview}
\begin{figure}[ht]
\vspace{.3in}
{\includegraphics[width=1\textwidth]{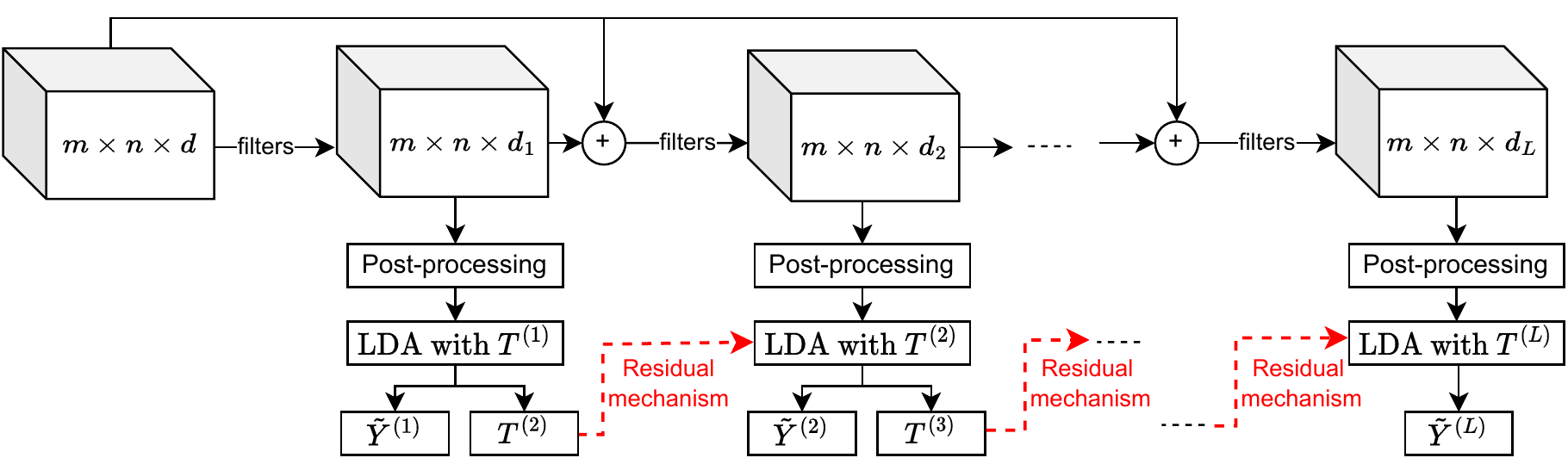}}%
\vspace{.3in}
\caption{The deep residual compensation convolutional network architecture. $T^{(1)}$ represents the original classes.}
\label{figd1}
\end{figure}

The network architecture, as shown in Figure \ref{figd1}, consists of multiple convolutional layers, each followed by post-processing steps and a linear discriminant analysis (LDA) classifier. Each convolutional layer, except the first layer, receives as input the concatenation of its previous layer's outputs and the original features. After post-processing the feature maps of the first layer, an LDA classifier is trained to categorise the extracted features using the original classes. By contrast, the LDA classifiers of the subsequent layers, referred to as compensation layers, are trained using new classes learnt from the residual information of the preceding layers. In fact, each layer produces two outputs ($\tilde{Y}^{(i)}$ and $T^{(i+1)}$ in Figure \ref{figd1}, where $i$ represents the $i^{th}$ layer). The first provides the predicted probabilities of the network at that layer, whereas the second represents the new classes produced for training the subsequent layer. To produce the network's outputs in each layer ($\tilde{Y}^{(i)}$ in Figure \ref{figd1}), the predicted probabilities of that layer are added to or subtracted from those of its preceding layers. This combination mechanism maintains the network's predicted probabilities of being in the range of between 0 and 1 and reduces the error of preceding layers. Section \ref{nc} describes in detail the network's components, including the convolutional layers (Section \ref{c1}), post-processing procedures (Section \ref{c2}) and residual mechanism (Section \ref{c3}). 
\subsection{Network Components}
\label{nc}
\subsubsection{Convolutional Layers} 
\label{c1}
Let $X^{(1)} \in \mathbb{R}^{m\times n\times d\times N}$ represent $N$ training images, $T$ is their $C$ classes, and $O^{(i)}\in \mathbb{R}^{m\times n\times d_i \times N}$ represents the feature maps produced by the $i^{th}$ layer. The $i+1$ layer receives as input the output of the $i^{th}$ layer concatenated with the original features; this input is represented as $X^{(i+1)}\in \mathbb{R}^{m\times n \times (d+d_i) \times N}$. We then divide the input images ($X^{(i+1)}$) into patches of $k\times k$ size and a stride of 1 pixel, where $k$ is the filter size and is a user-predefined parameter. The resulting matrix is $P \in \mathbb{R}^{k^2(d_i+d) \times \tilde{m}\tilde{n}N}$, where $\tilde{m}=(m-k)+1$, $\tilde{n}=(n-k)+1$ and $m$ and $n$ are the width and the height of the images, respectively. The filter learning process relies on applying any non-gradient-based method to the extracted local patches $P$ and collecting their weights to represent the filters used in the $i+1$ layer. The PCA filter bank by \citet{ref39} is an example of such filters. The authors first centralised the extracted patches to their mean to obtain $\bar{P} \in \mathbb{R}^{k^2(d_i+d) \times \tilde{m}\tilde{n}N}$. They then applied PCA to the centralised patches, where the principal components of ${\bar P}{\bar P^{T}}$ can be computed by solving the following optimisation problem: 
\begin{equation}
\begin{aligned}
    \min_{V\in R^{(k^2)\times (d_{i}+d)}} || \bar P - VV^T\bar P||_F^2,\\ s.t. \; \;V^TV=I, 
\end{aligned}
\end{equation}
where $I$ represents the identity matrix. The convolutional output of the $i+1$ layer can then be expressed as follows: 
\begin{equation}
    O^{(i+1)}= X^{i+1}\ast W^{(i)} \in \mathbb{R}^{m\times n\times d_{i+1}},
\end{equation}
where $d_{i+1}$ represents the number of filters in layer $i+1$, $X^{(i+1)}$ is the input images that are zero padded to obtain the same image size outputs, $\ast$ represents the convolution operation, and $W^{(i)}$ denotes the PCA filters expressed as follows: 
\begin{equation}
    \begin {aligned}
W^{(i)}= \underset{k\times k\times (d_{i}+d)}{\operatorname{mat}} q_s ,\;
s=1,2,\ldots,d_{i+1}, 
\end{aligned}
\end{equation}
where $q_s$  is the $s^\textrm{th}$ principal eigenvector of $\bar P{\bar P^T}$. 

The stacked-LDA filter bank is another example of filters generated without relying on gradient descent or backpropagation. Appendix \ref{AppendixA} discusses these filters in detail. These filters are computed using an iterative process that involves selecting a subset of the localised patches $P$ and then applying an LDA classifier to train the selected patches with their classes. The algorithm searches for patches with separable classes and accumulates their weights, which are subsequently used as stacked-LDA filters.

This article focuses mainly on the network architecture rather than investigating the filter type used. In our experiments (Section \ref{exp}), we use semi-supervised stacked-LDA filters created by combining 50\% of the supervised stacked-LDA filters with 50\% of the unsupervised PCA filters. However, different filter types can be employed, such as those used by \citet{ng2015dctnet}, \citet{yang2017canonical} and \citet{gatto2020semi}.\\
\subsubsection{Post-processing Steps}
\label{c2}
The feature maps of each convolutional layer are post-processed using a ReLU non-linear activation function, followed by second-order pooling and a multi-level spatial pyramid pooling. The ReLU function is applied to the feature maps of each layer but not between the layers. The feature maps are then pooled locally using the second-order pooling mechanism described by \citet{alotaibi2021multi}. Let $X_j^{(i)}\in \mathbb{R}^{m\times n\times d_i}$ denote the $j^{th}$ activations map in the $i^{th}$ layer, where $m$ and $n$ are the spatial dimensions of the images, and $d_i$ represents the number of filters in layer $i$. The calculation of the second-order pooling starts by dividing the tensors of $X_j^{(i)}$ into patches of the same size, which could be overlapped, e.g. $(r \times c)$. Each of these patches is then normalised using the z-score method, defined as $z=\frac{x-\mu}{\sigma}$, where $\mu$ and $\sigma$ represent the mean and the standard deviation of the data. Next, the channel-wise covariance matrix of each patch, after reshaping each of them to $rc \times d_i$, is computed. Because of the symmetry property of the covariance matrix, the number of second-order features is calculated as the number of patches$\times (\frac{d_i\times d_i}{2}+\frac{d_i}{2})$. The multi-level spatial pooling is then used to pool the second-order features. The multi-level spatial pyramid pooling calculation is identical to that implemented by \citet{chan2015pcanet}. Again, different post-processing procedures can be utilised; however, in our experiments (Section \ref{exp}), we found these steps to be the most effective for achieving high accuracy in the databases we used.
\subsubsection{Residual Mechanism}
\label{c3}
This mechanism is non-iterative; we add layers sequentially. Each layer is trained in a single pass with new labels learnt from the residual information of all its previous layers. The first layer's features are classified using an LDA classifier trained with the original classes. To produce the first layer's posteriors ($\tilde{Y}^{(1)}$ in Figure \ref{figd1}), we use the following sigmoid function on the output of the LDA classifier:
\begin{equation}
\label{sig}
    S(x)=\frac{1}{1+e^{-x/\sigma}},
\end{equation}
where $\sigma$ is the sigmoid scale parameter. To identify the new labels required to train the second layer ($T^{(2)}$ in Figure \ref{figd1}), we first find the residual errors between the current layer's predicted outputs ($\tilde{Y}^{(1)}$) and the original classes ($Y \in \mathbb{R}^{N\times C}$) in one-hot encoding, as follows: 
\begin{equation}
\label{equ1}
    R^{(1)}=\lambda Y-\tilde{Y}^{(1)}, 
\end{equation}
where $0 \leq \lambda \leq 1$ controls the maximum likelihood a class may attain. $T^{(2)}$ can then be defined as the classes with the maximum absolute residual errors, as follows:
\begin{equation}
\begin{aligned}
    T_i^{(2)}= &\class({|R_{ij}^{(1)}|,\forall j})&\;& i=[1,\dots,N],
    \end{aligned}
\end{equation}
where $\class(x)$ denotes the name of the class whose value has the largest residual error magnitude. Since our network is developed primarily for classification tasks, it concentrates on labelling rather than regressing the correction of posteriors. 

The second layer, as shown in Figure \ref{figd1}, receives an input $X^{(2)}\in \mathbb{R}^{m\times n \times (d+d_1) \times N}$ that is a concatenation of the first layer's feature maps and the original images. After finding the second layer's features, our objective is to learn a correction term based on the second layer classification results, which is then added to the posteriors of the first layer ($\tilde{Y}^{(1)}$) to provide more accurate predictions ($\tilde{Y}^{(2)}$). The correction term can either be positive or negative to maintain the network probabilities at the second layer ($\tilde{Y}^{(2)}$) to have values in the range of 0 to $\lambda$ (Equation \ref{equ1}). In other words, after training the second layer's LDA classifier using the new classes ($T^{(2)}$), the posteriors acquired can be added to or subtracted from the first layer's posteriors to correct them. The indicator variable ($s^{(2)}$) indicates whether to add to or subtract from the first layer's posteriors and is defined as the signum function of the maximum absolute residual errors of the previous layers, as follows:
\begin{equation}
\begin{aligned}
    s_i^{(2)}=&\sign(R_{i^*}^{(1)}), \, &i^*=\argmax_i|R_i^{(1)}|.
    \end{aligned}
\end{equation}
When the indicator values are positive, it indicates that the probabilities of the first and second layers are added, and when they are negative, the probabilities of the second layer are subtracted from those of the first. Consider a classification task of three classes $A$, $B$ and $C$. Assume that, given a single image whose actual class is $A$, the predicted probabilities of the first layer using this image are $0.4$, $0.6$ and $0$ for the three classes $A$, $B$ and $C$, respectively. To add a second layer, the residual error using $\lambda=0.8$ (Equation \ref{equ1}) is computed as $0.4$, $-0.6$ and $0$ for each of the three classes. Hence, the new label to train the second layer for this image is $B$, as it has the largest magnitude, and the indicator is $-1$ because the maximum absolute residual error has a negative sign. Assume that the second layer's features are trained using class $B$ and provided perfect prediction with probabilities of $0$, $1$ and $0$ for the three classes. Because the indicator value is negative, we subtract them from the predicted outputs of the first layer, resulting in $0.4$, $-0.4$ and $0$ for the three classes. Consequently, we reduce the error, and the predicted class of the second layer is $A$, which corresponds to the actual class of the image. To implement that and generalise it for the test set, as we do not know the classes, we divide the database into positive and negative samples based on their indicator variable values. The positive samples have positive indicator values, whereas the negative samples have negative indicator values. We then train two LDA classifiers for each layer; one is trained on the negative samples using $\{T_n^{(2)}\subset T^{(2)}: s^{(2)}=-1\}$, and the other is trained with the positive samples and their classes $\{T_p^{(2)}\subset T^{(2)}: s^{(2)}=1\}$. The negative classifier is trained by assuming that each class in the positive samples is a negative class (its class is zero). Similarly, during the training of the positive classifier, each class in the negative samples is treated as a negative class. Thus, both classifiers have access to all training data. The network's outputs at the second layer ($\tilde{Y}^{(2)}$) can then be expressed as follows:
\begin{equation}
    \tilde{Y}^{(2)}=\tilde{Y}^{(1)}+\alpha [\frac{n_p}{N}\,\tilde{Y_p}^{(2)}-\frac{n_n}{N}\,\tilde{Y_n}^{(2)}],
\end{equation}
where $\tilde{Y_n}^{(2)}$ and $\tilde{Y_p}^{(2)}$ are the $N$ predictions made by the classifiers trained on negative and positive samples, $n_n$ and $n_p$ denote the number of negative and positive examples, respectively, $N$ is the total number of training samples and $\alpha$ is a learning rate. The learning rate, similar to that used in neural networks, is introduced to reduce oscillations and provide faster convergence. However, the learning rate in our network also acts as a weight to integrate the probabilities of multiple layers, similar to weighted sum techniques. 

To add more layers, we repeat the steps used to add the second one. Algorithm \ref{algd1} summarises the procedures of the network's training with $L$ layers, assuming that the second-order features are known. To add layer $i$, the new labels $T^{(i)}$ and indicator variable ($s^{(i)}$) are computed based on the previous layer's residual error $R^{(i-1)}$, as shown in Equation \ref{eqd2}. The network's output at that layer can then be defined using Equation \ref{d_alpha}. In general, for deeper residual compensation layers, the new classes ($T^{(L)}$) learnt from the residual errors of the previous layers can be defined as follows: 
 \begin{equation}
 \label{equ3}
     \begin{aligned}
     T^{(L)}&=\class (|R^{(L-1)}|)\\
     &=\class(|Y-\tilde{Y}^{(L-1)}|)\\
     &=\class(|Y-[\tilde{Y}^{L-2}+\frac{\alpha}{N}(n_p^{(L-2)}
     \tilde{Y}_p^{(L-2)}-n_n^{(L-2)}\tilde{Y}_n^{(L-2)})])\\
     &=\vdots&\\
     &=\class(|Y-[\tilde{Y}^{(1)}+\frac{\alpha}{N}\sum_{i=2}^{L-1}(n_p^{i}\tilde{Y}_p^{i}-n_n^{i}\tilde{Y}_n^{i})]),
     \end{aligned}
 \end{equation}
where $\class(x)$ is the name of the class whose value has the largest residual error magnitude, $Y$ is the original classes in one-hot encoding, $n_p^{i}$ and $n_n^{i}$ are the number of positive and negative samples in layer $i$, respectively, and $N$ denotes the total number of samples in the database.
%########################################################
\begin{algorithm}[ht]
\caption{Deep Residual Compensation Convolutional Network Training}\label{algd1}
\begin{algorithmic}[1]
%Input and Output
\REQUIRE Second-order features: $\{F^{(i)},\;i=[1,2,\dots,L]\}$, $L$ number of layers, $C$ classes $T^{(1)}\in \mathbb{R}^{N\times 1}$, learning rate: $\alpha$ and $\lambda$ to determine the highest probability a class can reach. 
\ENSURE Model's accuracy: $accuracy$
% assign with arrow
\STATE Generate $Y\in \mathbb{R}^{N\times C}$, the one-hot encoding of $T^{(1)}$. 
\STATE $i$ $\gets$ $1$ and $\tilde{Y}^{(0)} \gets 0$.
\STATE $s^{(i)}=1^{N\times 1}$ \COMMENT{fill the first layer's indicator variable with 1. }
\WHILE{$i<L$}
\IF{$i>1$} 
\STATE { Find the residual errors, new classes and indicator variable, as follows:}
\begin{equation}
\label{eqd2}
\begin{aligned}
    &R^{(i-1)}=\lambda Y-\tilde{Y}^{(i-1)}. \\
    &T_j^{(i)}=\class({|R_{jk}^{(i-1)}|,\forall k}), \\
    &  s_{j}^{(i)}=\sign(R_{j^{\ast}}^{(i-1)}), \, j^*=\argmax_j|R_{j}^{(i-1)}|,\\
    & j=[1,2,\dots,N].
\end{aligned}
\end{equation}
\ENDIF
\STATE { Find $F_n^{(i)}\subset F^{(i)}$ and $T_n^{(i)}\subset T^{(i)}$, for which their indicator values are negative.}  
\STATE { Find $F_p^{(i)}\subset F^{(i)}$ and $T_p^{(i)}\subset T^{(i)}$, for which their indicator values are positive.}  
\IF{$T_p^{(i)} \neq \emptyset$}
\STATE $L_1=\LDA(F_p^{(i)},T_p^{(i)})$. 
\STATE $\tilde{Y}_p^{(i)}=\prediction(L_1,F^{(i)}).$
\ELSE
\STATE $\tilde{Y}_p^{(i)} \gets 0$.
\ENDIF
\IF {$T_n^{(i)} \neq \emptyset$}
\STATE $L_2=\LDA(F_n^{(i)},T_n^{(i)})$. 
\STATE  $\tilde{Y}_n^{(i)}=\prediction(L_2,F^{(i)}).$
\ELSE 
\STATE  $\tilde{Y}_n^{(i)} \gets 0$.
\ENDIF
\STATE {Compute the current's layer output $\tilde{Y}^{(i)}$, as
\begin{equation}
\label{d_alpha}
    \tilde{Y}^{(i)}=\tilde{Y}^{(i-1)}+\alpha [\frac{n_p}{N}\,\tilde{Y_p}^{(i)}-\frac{n_n}{N}\,\tilde{Y_n}^{(i)}],
\end{equation}
where $n_p$ and $n_n$ represent the number of positive and negative samples, respectively. }
\STATE $i\gets i+1$
\ENDWHILE
\STATE  Compute the model's accuracy at layer $L$ using $\tilde{Y}^{(L)}$.
\end{algorithmic}
\end{algorithm}
\section{EXPERIMENTS}
\label{exp}
\subsection{Databases}
\label{data}
We used four standard benchmarks in our experiments: CIFAR-10 \citep{ref_cifar}, CIFAR-100 \citep{ref_cifar}, MNIST \citep{ref30} and TinyImageNet \citep{ref_tiny}. The MNIST database comprises 60,000 training examples and 10,000 test images of size $28\times 28$, drawn from the same distribution, normalised and centred in a fixed-size image. The CIFAR-10 database consists of 10 classes with 50,000 images for training and 10,000 test images. The images of size $32\times 32\times 3$ have a low resolution with different poses and angles. CIFAR-100 is similar to CIFAR-10 but with 100 classes. The TinyImageNet database consists of 100,000 training images of size $64 \times 64 \times  3$. The images are divided into 200 categories, with 500 images each. The validation and the test sets contain 10,000 images each, with 50 images per class. The test set is not labelled, and our experiments’ performance is reported on the validation set. 
\subsection{Image Classification without Data Augmentation}
\label{s1}
In this section, we evaluate the proposed network on the MNIST, CIFAR-10, CIFAR-100 and TinyImageNet databases (Section \ref{data}) without data augmentation. To determine the network parameters, we examine a variety of configurations, each of which has different parameters, and report the results of the configuration that works the best. 
\subsubsection{Parameter Settings}
\label{setting}
The optimal settings identified for the MNIST, CIFAR-10, CIFAR-100 and TinyImageNet databases are listed in Table \ref{tabd2}. All architectures, except the MNIST database, used $3 \times 3$-pixel filters created by combining PCA \citep{ref39} and stacked-LDA filters (Appendix \ref{AppendixA}) at a 50\% ratio. The MNIST database used $13 \times 13$ PCA filters in its first layer and $3\times 3$ PCA filters in its residual layers. In this section and throughout the rest of this article, we used the same number of filters for all layers and stopped adding layers when the training error rate approached 0\%. Therefore, the number of filters per layer is 60 in the MNIST's architecture, 50 in the CIFAR's architectures and 40 in TinyImageNet's architecture. The number of layers beyond which an accuracy gain was no longer observed was 937 for the CIFAR-10 database, 436 for the CIFAR-100 database, 231 for the MNIST database and 512 for the TinyImageNet database. We utilised $7\times 7$-block second-order pooling for MNIST with a stride of four pixels, $16\times 16$ for CIFAR-100 with a four-pixel stride, $16\times 16$ for CIFAR-10 with a one-pixel stride, and $32\times 32$ for TinyImageNet with an eight-pixel stride. In all architectures, we pooled the second-order features using three-level spatial pyramid pooling of $4\times 4$, $2\times 2$ and $1\times 1$ subregions. The only data preprocessing was min-max normalisation applied to the input of each convolutional layer, and probabilities were retrieved from all datasets except MNIST using the sigmoid function with a scale value of 16 (Equation \ref{sig}). In the MNIST database, we used the following softmax function to generate the probabilities in each layer: 
\begin{equation}
    \softmax(y_i)=\frac{\exp(\beta y_i)}{\sum_{j=1}^C \exp(\beta y_j)},
\end{equation}
where $\beta$ is assigned to 0.001, $C$ denotes the number of classes, and $y$ represents the outputs of the LDA classifier. During the training phase, $\lambda$ (Equation \ref{equ1}) was set to $0.8$, and the learning rate was fixed at $\alpha=1$ for the MNIST database and $\alpha=0.4$ for the CIFAR-10 database. The remaining databases used an initial learning rate of 1, which was dropped by 10\% every 10 layers as follows:
\begin{equation}
\label{lr_red}
    \alpha=\alpha - \frac{10}{100} \alpha.
\end{equation}
We stopped reducing the learning rate when it reached 0.387 and 0.478 for CIFAR-100 and TinyImageNet databases, respectively. 
\begin{table} [htbp]
\caption{Network architectures using the MNIST, CIFAR-10, CIFAR-100 and TinyImageNet databases}
\label{tabd2}
\centering
\begin{tabular}{|c|c|c|}
\hline
\multicolumn{3}{|c|}{The MNIST database: $28\times 28\times 1$}\\
\hline
\textbf{Filter size}& \textbf{SOP}&\textbf{Output size}\\
\hline
$13 \times 13 \times 1\times 60$& $7\times 7$, Stride $=4$&$28\times 28\times 60$\\
$[3 \times 3 \times 60\times 60]\times 230$& $7\times 7$, Stride $=4$&$28\times 28\times 60$\\
\hline
\multicolumn{3}{|c|}{The CIFAR-10 database: $32\times 32\times 3$}\\
\hline
\textbf{Filter size}& \textbf{SOP}&\textbf{Output size}\\
\hline
$3 \times 3 \times 3\times 50$& $16\times 16$, Stride $=1$&$32\times 32\times 50$\\
$[3 \times 3 \times 50\times 50]\times 936$& $16\times 16$, Stride $=1$&$32\times 32\times 50$\\
\hline
\multicolumn{3}{|c|}{The CIFAR-100 database: $32\times 32\times 3$}\\
\hline
\textbf{Filter size}& \textbf{SOP}&\textbf{Output size}\\
\hline
$3 \times 3 \times 3\times 50$& $16\times 16$, Stride $=4$&$32\times 32\times 50$\\
$[3 \times 3 \times 50\times 50]\times 435$& $16\times 16$, Stride $=4$&$32\times 32\times 50$\\
\hline
\multicolumn{3}{|c|}{The TinyImageNet database: $64\times 64\times 3$}\\
\hline
\textbf{Filter size}& \textbf{SOP}&\textbf{Output size}\\
\hline
$3 \times 3 \times 3\times 40$& $32\times 32$, Stride $=8$&$64\times 64\times 40$\\
$[3 \times 3 \times 40\times 40]\times 511$& $32\times 32$, Stride $=8$&$64\times 64\times 40$\\
\hline
\end{tabular}
\end{table}
%########################################################
\subsubsection{Performance Analysis}
Table \ref{tabd4} reports the accuracy of ResCNet compared with some gradient-based and non-gradient-based networks on the MNIST, CIFAR-10, CIFAR-100 and TinyImageNet databases without data augmentation. The non-gradient-based models reported in Table \ref{tabd4} are the best-performing PCANet-like models from the perspective of the datasets we considered. Such networks include PCANet \citep{chan2015pcanet}, LDANet \citep{chan2015pcanet}, DFSNet \citep{gatto2020semi} and Multi-layer PCANet \citep{alotaibi2021multi}. To show where our network fits in the literature of commonly used networks, we compared the performance achieved by our network to those obtained by standard gradient-based convolutional networks. Maxout \citep{goodfellow2013maxout}, Network in network \citep{lin2013network}, stochastic pooling \citep{zeiler2013stochastic} and ResNet \citep{ref4} are the gradient-based networks listed in the table. The accuracy of 164-ResNet with pre-activation was reported by \citet{huang2017densely}, while the results of ResNet-18 and ResNet-34 were reported by \citet{jeevan2022convolutional}.

According to Table \ref{tabd4}, for the MNIST database, our model with 231 layers achieved an accuracy of 99.52\%, making it superior to all non-gradient-based models, such as PCANet, Multi-Layer PCANet and LDANet, in terms of accuracy. It improved on the best results of the non-gradient-based networks by roughly 0.12\%. In addition, the findings presented in the table demonstrated that our network produced results comparable with standard gradient-based networks, such as the Maxout network, Network in network and stochastic pooling. 

As shown in Table \ref{tabd4}, for the CIFAR-100 database, the accuracy attained by our proposed network was the highest among all the networks. The accuracy of 64.91\% was around 8\% higher than that of Multi-Layer PCANet and stochastic pooling, more than 9\% higher than that of ResNet-110 and ResNet-32, 14\% higher than the original PCANet and more than 2\% higher than that of Maxout and ResNet with stochastic depth. Moreover, the results achieved by our network were roughly equivalent to those obtained by ResNet with 164 layers and Network in network using the dropout technique. 

For TinyImageNet, as shown in Table \ref{tabd4}, our network achieved the highest accuracy among all the networks without any data augmentation. This performance was around 2\% higher than that of the ResNet-34 model and 1\% better than that of the ResNet-18 model. 

 According to the results shown in Table \ref{tabd4} for the CIFAR-10 database, our model outperformed PCANet and all of its variants, in terms of accuracy. The accuracy of the proposed network was around 10\% better than the performance achieved by the original PCANet and 6\% higher than that of Multi-Layer PCANet. Although ResCNet’s accuracy was around 1\% lower than that of the Maxout network and 2\% worse than that of Network in network, our model achieved accuracy comparable with that of ResNet-32, 1\% higher than ResNet-18 and ResNet-110, and 3\% greater than stochastic pooling. In general, ResCNet with more than 900 layers showed an excellent performance of 87.54\% on the CIFAR-10 database with no data augmentation, making it the first PCANet-like network to reach such a number of layers and such a performance.
 
In general, Table \ref{tabd4} demonstrates that ResCNet outperformed all PCANet-like networks in terms of accuracy and the number of layers required for training. It also shows that our model, which is trained without complicated non-linear functions or regularisation techniques, achieves accuracy similar to that of standard convolutional networks such as Network-in-Network, Maxout, stochastic pooling, and several residual networks. Even though the number of layers in our network is relatively large, these layers are added sequentially, one after the other, without iterations or intensive parameter updates. In addition, we used a small number of filters in each layer for all of our architectures, with no configuration exceeding 60 filters per layer. To the best of our knowledge, ResCNet is the first non-gradient-based propagation-free network to be trained with hundreds of layers. 
\begin{table} [htbp]
\centering 
\caption{Accuracy (\%) of ResCNet compared with different methods on the CIFAR-10 (C10), CIFAR-100 (C100), MNIST and TinyImageNet (T200) databases without data augmentation  }
\label{tabd4}
\begin{tabular}{|c|c|c|c|c|}
\hline 
\multicolumn{5}{|c|}{\emph{Non-gradient-based networks}}\\
\hline
\textbf{Method}&\textbf{C10}&\textbf{C100}&\textbf{MNIST}&\textbf{T200}\\
\hline
\makecell{PCANet-2} & 77.14&51.62&99.34&30\\
\makecell{LDANet} &78.33&--&99.38&--\\
\makecell{DFSNet-3} &81.06&--&--&--\\
\makecell{Multi-Layer PCANet} &81.72&57.86&99.40&40.87\\
\makecell{ResCNet (ours)}&  \text{87.54}&\textbf{64.9}&{99.52}&\textbf{44.37}\\
\hline
\multicolumn{5}{|c|}{\emph{Gradient-based networks}}\\
\hline
\textbf{Method}&\textbf{C10}&\textbf{C100}&\textbf{MNIST}&\textbf{T200}\\
\hline
Stochastic pooling &84.87&57.49&99.53&--\\
\makecell{Maxout network, with dropout} &\text{88.32}&61.43&\textbf{99.55}&--\\
\makecell{Network in network, with dropout} &\textbf{89.59}&64.32&99.53&--\\
\makecell{Network in network, without dropout} &\text{85.49}&--&--&--\\
\makecell{110 ResNet} & 86.82&55.26&--&--\\
\makecell{ResNet stochastic depth} &-&62.20&--&--\\
\makecell{164-ResNet (pre-activation)} &-& {64.42}&--&--\\
\makecell{ResNet-18} &86.29& 59.15&--&43.02\\
\makecell{ResNet-32} &87.97& 56.05&--&42.65\\
%resnet reported by [18] in the previous chapter 
\hline
\end{tabular} 
\end{table}
%###############################################################################

\subsection{Network Parameters}
\label{ss2}
This section explains how to configure ResCNet's parameters by analysing their impact on the model's accuracy using the CIFAR-10 database. The experiments in this section are divided into two subsections, as follows: 
\begin{itemize}
    \item \textbf{Experiment 1:} Testing the effect of the number of filters on the accuracy of ResCNet
    \item \textbf{Experiment 2:} Studying the impact of learning rate on the model's performance
\end{itemize}
\subsubsection{Number of Filters}
\label{experiment1}
This experiment aimed to determine how changing the number of filters affects the accuracy of ResCNet. For this, we designed two networks with the same settings but different numbers of filters. The first, referred to as ResCNet-30, used 30 filters in each of its layers, while the second (ResCNet-50) used 50 filters in all of its layers. The two networks followed the same parameters settings as those described in Section \ref{s1} for the CIFAR-10 database. However, we used an 8x8 second-order pooling block size. 

\begin{figure}[h]
\vspace{.3in}
\centerline{\includegraphics[width=1\textwidth]{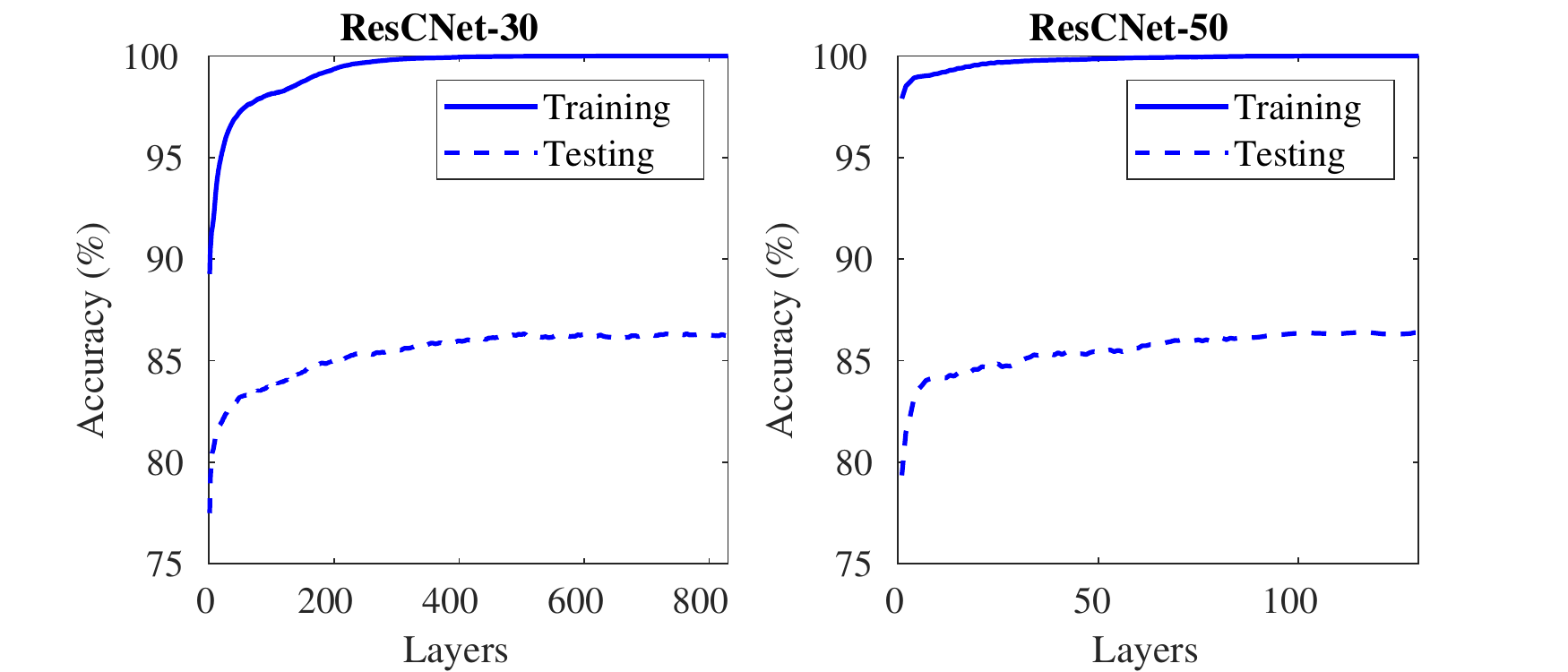}}
\vspace{.3in}
\caption{The accuracy (\%) of ResCNet-30 and ResCNet-50 on the training and testing sets of the CIFAR-10 database with no data augmentation. }
\label{figd2}
\end{figure}

Figure \ref{figd2} shows the accuracy of training and testing ResCNet-30 and ResCNet-50 on the CIFAR-10 database without data augmentation. According to the figure, both networks achieved the same level of accuracy, although ResCNet-30 required more layers. ResCNet-50 obtained a training accuracy of 100\% at layer 127, while ResCNet-30 achieved 100\% at layer 781. The testing accuracy for both 127- and 781-layer networks was around $86.28\%$. The best testing accuracy of $86.33\%$ was obtained with 100-layer ResCNet-50 and 500-layer ResCNet-30. The findings in this section suggested that, unlike previous PCANet-like networks in which the number of filters in each layer should be determined in advance, we could obtain the same performance by employing any number of filters based on the available resources. However, we needed more filters in each layer to achieve the desired performance faster. The results in this section also showed that the second-order pooling block size affected the accuracy of ResCNet. For instance, the highest accuracy reached for the CIFAR-10 database (87.54\% in Section \ref{s1}) was around 1\% better than that achieved in this section (86.33\%), with the only difference between the models being the modification of the second-order pooling block size.
%##############################################
\subsubsection{Learning Rate}
\label{expeiment2}
This experiment aimed to show the impact of the learning rate ($\alpha$ in Equation \ref{equ3}) on ResCNet's accuracy. Similar to neural networks, the learning rate is introduced to prevent oscillations and promote faster convergence. We designed two networks with the same parameters but different learning rates and evaluated them on the CIFAR-10 database. Both networks used 50 filters in all of their layers. The first network (ResCNet-50--1) was trained with a learning rate of $\alpha=1$, whereas the second (ResCNet-50-0.4) was trained using $\alpha=0.4$. The other network's parameters were the same as those described in Section \ref{s1} for the CIFAR-10 database.
% ref{tab11}

Table \ref{tab11} compares the accuracy of ResCNet-50--1 with that of ResCNet-50--0.4 using a different number of layers on the CIFAR-10 database. The training accuracy of ResCNet-50--1 reached 100\% at layer 384, with a testing accuracy of 86.91\%. On the other hand, ResCNet-50--0.4 obtained a 100\% training accuracy at layer 970, with a testing accuracy of 87.41\%. The optimal testing accuracy of ResCNet-50--1 was achieved at layer 208, with 87.2\% testing accuracy, while ResCNet-50--0.4 obtained its best performance of 87.54\% at layer 937. The results achieved in this section demonstrated that the learning rate $\alpha= 0.4$ was small, as we needed more than 600 layers for the network to reach its convergence. If the training error rate oscillates, which is not the case in CIFAR-10, the learning rate must be decreased to avoid oscillations. 
\begin{table} [htbp]
\caption{Accuracy (\%) on the CIFAR-10 database test set using ResCNet-50--1 and ResCNet-50--0.4 }
\label{tab11}
\centering
\begin{tabular}{|c|c|c|}
\hline
\textbf{Network}&\textbf{\# Layers}&\textbf{Accuracy (\%)}\\
\hline 
ResCNet-50--1&348&86.91\\
ResCNet-50--1&208&87.02\\
\hline
ResCNet-50--0.4&970&87.41\\
ResCNet-50--0.4&937&\textbf{87.54}\\
ResCNet-50--0.4&600&87.09\\
\hline
\end{tabular}
\end{table}
\subsection{Image Classification with Data Augmentation}
\label{s2}
This section aims to enhance the classification accuracy of ResCNet using data augmentation on three databases, namely, CIFAR-10, CIFAR-100 and TinyImageNet. The ResCNet architectures in this section shared the same design and parameters as those in Section \ref{s1} but had 490, 507 and 480 layers for the CIFAR-10, CIFAR-100 and TinyImageNet databases, respectively. For CIFAR-10, we modified the pooling stride to four pixels and re-implemented a 193-layer ResCNet without data augmentation for adequate comparison.

Table \ref{tabd8} compares the accuracy of ResCNet with and without data augmentation on the CIFAR-10, CIFAR-100 and TinyImageNet databases. With horizontal flipping being the only data augmentation used, the accuracy of ResCNet was enhanced across all three databases. This improvement was around 1\% for the TinyImageNet database, 2\% for the CIFAR-10 database and 3\% for the CIFAR-100 database. For the CIFAR-10 database, the accuracy achieved with data augmentation was also 1\% higher than the best result (87.54\%) reported in Section \ref{s1}. 

This section’s results showed the importance of data augmentation for improving the model’s generalisation and accuracy. Incorporating other data augmentation types are needed to increase the model’s accuracy further. Since ResCNet is not currently being trained in a batch-based manner, adding more data augmentation types is challenging as they need to be computed in advance. In future works, we will investigate the possibility of converting the current work into a batch-based system as in gradient-based models. 
 
\begin{table} [htbp]
\centering 
\caption{Accuracy (\%) of ResCNet on the CIFAR-10, CIFAR-100 and TinyImageNet databases }
\label{tabd8}
\begin{tabular}{|c|c|c|c|}
\hline 
\textbf{Data augmentation}&\textbf{CIFAR-10}&\textbf{CIFAR-100}&\textbf{TinyImageNet}\\
\hline
\xmark&86.82&64.9&44.37\\
\cmark&\textbf{88.35}&\textbf{67.8}&\textbf{45.91}\\
\hline
\end{tabular} 
\end{table}
\section{CONCLUSIONS}
\label{s3}
In this article, we proposed ResCNet, a PCANet-like network that trains each layer with new labels derived from the residual data of all preceding layers. Our proposed network increased the network depth to more than 950 layers, making it the first non-gradient-based propagation-free network to achieve this number. Moreover, ResCNet's performance was comparable to that of standard gradient-based models and superior to PCANet and all of its variants. Increasing the size of the databases by one type of data augmentation resulted in a considerable improvement in accuracy, particularly in CIFAR-100, where it reached 3\%. However, increasing the number of samples leads to higher computational costs. To overcome this issue and in future work, we will investigate the possibility of transforming the current network into a batch-based system, similar to the neural network. In addition, the network may be developed further by modifying the filter types or by using other metrics to define the residual errors and, thus, the new classes.
\bibliographystyle{abbrvnat}
\bibliography{references}  %%% Uncomment this line and comment

\newpage
\appendix
\section{STACKED-LDA}
\label{AppendixA}
In this appendix, we introduce the stacked-LDA method, a model that stacks two Linear discriminant analysis layers. The first layer of the model is trained using new labels that are produced by clustering similar instances of a certain class. The second layer, on the other hand, is trained using the original classes. This general concept of the stacked-LDA model is described in Section \ref{c11}. We propose an iterative method to create clusters of similar groups. The description of our method is explained in Section \ref{c21}. In addition, Section \ref{c31} discusses the procedures for using the stacked-LDA algorithm as convolution filters. Finally, the parameter settings we used in our experiments in the main paper were described in Section \ref{c41}.
\subsection{General Concept}
\label{c11}
The stacked-LDA algorithm relies on stacking two linear discriminant analysis (LDA) layers. The first LDA layer is trained using labels created from the original classes, while the second LDA layer is trained using the actual labels. To generate the classes of the first layer, similar samples from a given class are grouped to create a new class. For example, a class $A$ with $s$ instances can be subdivided into $c$ new classes, each with a different number of instances. The minimum number of samples needed to represent a class is one. Suppose the first LDA can differentiate between the new classes ideally. In that case, the second LDA will receive the posteriors of the first LDA and be able to differentiate between the actual classes. For instance, if class $A$ is subdivided into $A_1$ and $A_2$, the subsequent LDA will identify that both $A_1$ and $A_2$ are members of class $A$. 
\subsection{Stacked-LDA Algorithm}
\label{c21}
Let $X\in \mathbb{R}^{N\times M}$ represent $N$ training samples, each with $M$ dimensions, and $Target \in \mathbb{R}^{N\times 1}$ represent their original classes. Algorithm \ref{algs3} describes the procedures for applying the stacked-LDA to the training set. The algorithm starts by picking a random class $c$ from the actual classes. We then choose $N_{positive}$ random instances that belong to class $c$ and $N_{negative}$ random examples that are not in class $c$, where $N_{positive}$ and $N_{negative}$ are the number of positive and negative samples and are user-predefined parameters. Next, an LDA classifier discriminates between the positive and the negative samples. After that, we check if our chosen random samples are linearly separable, which can be done by comparing the error rate of the LDA classifier with a small value of nearly zero called tolerance ($tol$) and is chosen by the user. If the LDA's error rate is lower than the tolerance, we consider the positive class to be a new class and collect the weights of the LDA. On the other hand, if the LDA's error rate is greater than the tolerance, the chosen samples are not similar and cannot be grouped. Therefore, the algorithm proceeds to find other classes that separate the data accurately in the same way until reaching the required number of classes ($N\_classes$). When algorithm \ref{algs3} terminates, we can use the generated LDA's weights to find the output of the first LDA. Another LDA can then be applied to the output of the first LDA to classify them back using the original classes. 
\begin{algorithm}
\caption{Stacked-LDA Algorithm}\label{algs3}
\begin{algorithmic}[1]
%Input and Output
\REQUIRE Training set: $ X\in\mathbb{R}^{N\times M},$ where $\{x_i^N\; ,\; x_i\in \mathbb{R}^M\}$, original classes: $Target \in \mathbb{R}^{N\times 1}$, number of classes user wants to generate: $N\_classes$, number of positive samples: $N_{positives}$, number of negative samples: $N_{negatives}$ and tolerance of performance user can afford: $tol$
\ENSURE LDA's weights: $weights \in \mathbb{R}^ {N\times N\_classes}$ and LDA's bias or constant: $bias \in \mathbb{R}^{N\_classes}$
% assign with arrow
\STATE $weights \gets [\;]$.
\STATE $bias \gets [\;]$. 
\STATE $i \gets 1$. 
\WHILE{$i<N\_classes$}
\STATE Pick a random class $c$ from the Target. 
\STATE Pick random $N_{positives}$ samples from class c ($S_{positives}$). 
\STATE {Choose $N_{negatives}$ samples that are not in class c ($S_{negatives}$) randomly.}
\STATE Combine the negative and positive samples: $S \gets [S_{positives}, S_{negatives}]$. 
\STATE {$T \gets [ones(N_{positives}), zeros(N_{negatives})]$.} 
\STATE {Find the linear discriminant analysis (LDA) between $S$ and $T$: $L=\LDA(S,T)$.} 
\STATE Find the perfromance ($ErrorRate$) of $S$ using $L$. 
\IF{$ErrorRate<tol$}
\STATE $weights \gets [weights, LDA's weights]$. 
\STATE $bias \gets [bias, LDA's bias]$. 
\STATE $i \gets i+1$.
\ENDIF
\ENDWHILE
\end{algorithmic}
\end{algorithm}
\subsection{Stacked-LDA Filters}
\label{c31}
For $N$ training samples $X$: $\{X_i\in \mathbb{R}^{m \times n \times c}\}$ with actual classes
$Target \in \mathbb{R}^{N\times 1}$, where $m$ and $n$ are the spatial dimensions of the image and $c$ is the number of channels, we compute the stacked-LDA filters as follows: 

\begin{enumerate}
    \item Given a single image $X_i\in \mathbb{R}^{m\times n\times c}$ and a filter size $k_L \times k_L$, we extract and vectorise all overlapping patches of size $k_L\times k_L \times c$ each. The resulting matrix is $P_i\in \mathbb{R}^{(k_L^2 \times c)\times \tilde m \tilde n}$, where $\tilde{m}=(m-k_L)+1$, $\tilde{n}=(n-k_L)+1$ and $m$ and $n$ are the spatial dimensions of the image, respectively; 
    \item  We repeat the previous step for all images in the dataset to obtain $P\in \mathbb{R}^{(k_L^2 \times c)\times \tilde m \tilde n N}$;
    \item We create a vector $T \in \mathbb{R}^{1\times \tilde m  \tilde n N}$ that contains the class labels of the patches. A single patch is assigned a label equivalent to the class of its full image;
    \item  Using random samples and specific tolerance, we apply Algorithm \ref{algs3} on $P$ and $T$ to obtain the stacked-LDA filters' weights $W_s^L$ and bias $B_s^L$; 
    \item We can express the Stacked-LDA filters as follows: 
    \begin {equation}
\begin {split}
& W_s^L= \underset{k_L\times k_L\times c}{\operatorname{mat}} q_s ,\;
s=1,2,\ldots,d_L,\\
& B_s^L= \underset{1\times 1\times c}{\operatorname{mat}} q_s ,\;
s=1,2,\ldots,d_L,
\end{split}
\end{equation}
where $d_L$ is the number of filters chosen by the user, which is equivalent to the number of classes in Algorithm \ref{algs3};
\item We convolve the original images with the filters as follows: 
\begin{equation}
X_i^L=X_i^{L-1}\ast W_s^L+ B_s^L \in \mathbb{R}^{m\times n\times d_L},
\end{equation}
where $s=1,2, \ldots, d_L$ and $X_i^{L-1}$ is zero-padded to obtain the same image size; 
\end{enumerate}
\subsection{Parameter Settings}
\label{c41}
To compute the stacked-LDA filters in our experiments in the main paper, we set the number of positive samples ($N_{positives}$) to $2$, while the number of negative samples ($N_{negatives}$) was  32. The tolerance ($tol$) in Algorithm \ref{algs3} was set to zero. We used the LDA classifier with the one-versus-all decomposition method.
%%%%%%%%%%%%%%%%%%%%%%%%%%%%%%%%%%%%%%%%%%%%%%%%%%%%%%%%%%%%%%%%%%%%%%%%%%%%%%%
%%%%%%%%%%%%%%%%%%%%%%%%%%%%%%%%%%%%%%%%%%%%%%%%%%%%%%%%%%%%%%%%%%%%%%%%%%%%%%%
\end{document}